\pdfoutput=1

\documentclass[11pt]{article}

\usepackage[review]{ACL2023}

\usepackage{times}
\usepackage{latexsym}
\usepackage{soul}
\usepackage{tabularx}
\usepackage{graphicx} 
\usepackage{subcaption} 
\usepackage{longtable}
\usepackage{hyperref}
\usepackage[T1]{fontenc}

\usepackage[utf8]{inputenc}
\usepackage{microtype}
 \usepackage{amsmath}

\usepackage{inconsolata}
\usepackage{xcolor}
\usepackage{lineno}
\title{Mitigating Biases to Embrace Diversity: A Comprehensive Annotation Benchmark for Toxic Language}

\author{Xinmeng Hou \\ Columbia University \\
  \texttt{fh2450@tc.columbia.edu}}

\begin{document}
\maketitle
\nolinenumbers
\begin{abstract}
This study introduces a prescriptive annotation benchmark grounded in humanities research to ensure consistent, unbiased labeling of offensive language, particularly for casual and non-mainstream language uses. We contribute two newly annotated datasets that achieve higher inter-annotator agreement between human and language model (LLM) annotations compared to original datasets based on descriptive instructions. Our experiments show that LLMs can serve as effective alternatives when professional annotators are unavailable. Moreover, smaller models fine-tuned on multi-source LLM-annotated data outperform models trained on larger, single-source human-annotated datasets. These findings highlight the value of structured guidelines in reducing subjective variability, maintaining performance with limited data, and embracing language diversity.

\textbf{Content Warning}: This article only analyzes offensive language for academic purposes. Discretion is advised.
\end{abstract}

\section{Introduction}
To properly offer people the option to avoid potentially offensive language 
while also protecting minoritized language varieties 
from being misidentified, accurate detection that can identify languages despite changes over time is required. Current datasets typically employ multifaceted methodologies for content categorization, taking into account not just the presence of offensive language but also its context, target, and underlying intent \cite{zampieri2019predicting, basile2019semeval, mollas2020ethos}. Abusive, toxic, or offensive language and hate speech were often directly identified based on finite lists of phrases \cite{davidson2017automated}, 
annotators' interpretation of the textual content \cite{gibert2018hate, founta2018large, sap2019social, susanto2024indotoxic2024demographicallyenricheddatasethate}, or a combination of both \cite{vargas2021contextual, basile2019semeval}. 
This raises the issue of an unclear research subject characterized by inconsistencies in terminology and categorization \cite{fortuna2020toxic}. For instance, hate speech is often treated as equivalent to offensive or toxic language \cite{susanto2024indotoxic2024demographicallyenricheddatasethate}, which leads to problems where language that is less offensive than hate speech may be incorrectly classified as non-offensive.

Biases in annotation refer to the systematic tendency of human annotators that leads to errors or skewed labels in the training data used for machine learning models \cite{10.1162/tacl_a_00550}. The most common approach for mitigating annotator bias is diversifying annotation teams and increasing annotation on each raw piece \cite{10.1162/tacl_a_00550, sap2019social, Geva2019AreWM}. However, no research addresses how diverse the annotator team should be and how many annotators were required to eliminate bias efficiently. While diversification and scale help address bias, the root issue often lies in subtle differences in interpretations addressing complex socio-cultural dynamics that are especially vulnerable \cite{Kuwatly2020IdentifyingAM}. Therefore, rather than treating annotator disagreement as mere "noise" or using majority vote labels to cover up disagreement, inevitable disagreements should be adequately addressed in annotation \cite{10.1162/tacl_a_00550, Davani2021DealingWD}. The main research question is \textbf{how to reveal the underlying patterns while minimizing the impact of biased annotations against non-standard language use during the data labeling process to protect language diversity}. Moreover, data may be limited or nonexistent, particularly for endangered dialects, minority language use \cite{Liu2022NotAA}, and low-resource scenarios. The second question explores \textbf{whether annotated features can improve models' robustness against small datasets and varied language use, making them more accommodating of English variety}. Finally, we observed that skilled and well-trained human annotators are not always readily available. Instead of relying on untrained annotators who lack expertise in language or social studies, we investigate \textbf{whether prompted large language models (LLMs) can serve as a viable alternative}.

\begin{figure}[tb]
    \begin{center}
    \includegraphics[width=2.5in]{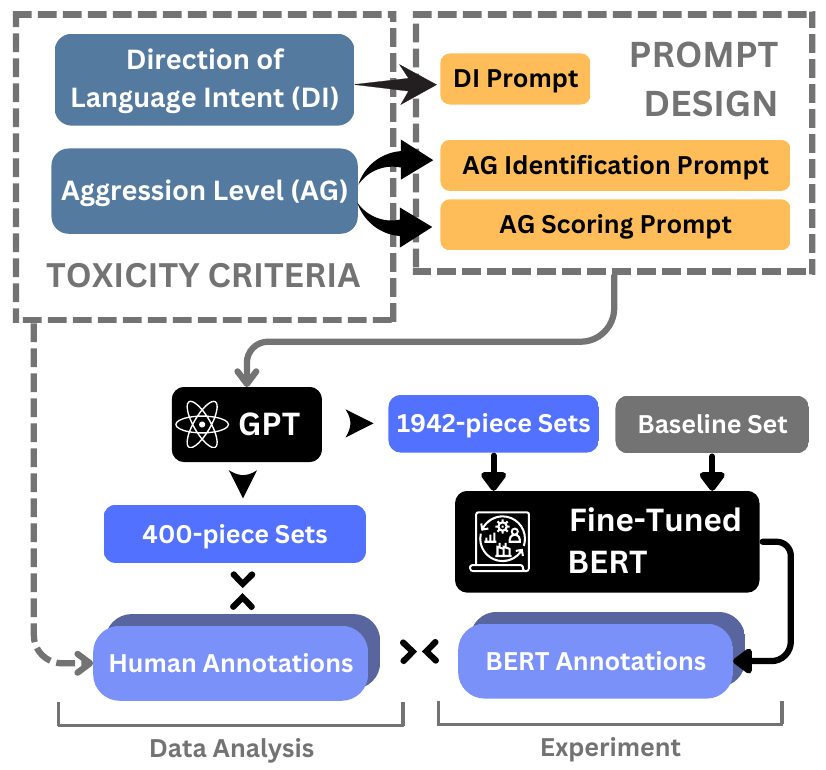}
    \caption{\textbf{Research Design}: This research establishes standardized criteria for toxic language annotation and analyzes inter-annotator reliability. Experiments on BERT models across language types tend to demonstrate the broader applicability of the proposed annotation criteria, even with limited resources.} 
    \label{fig:enter-label}
    \end{center}
\end{figure}

As shown in Figure \ref{fig:enter-label}, our research addresses three key components: (1) proposing criteria for a prescriptive annotation framework that will be introduced in methodology, (2) conducting a small-scale statistical analysis to compare the framework with the descriptive paradigm and evaluate the performance of prescriptively-prompted LLMs, and (3) testing the framework under limited conditions, using smaller datasets with complex language features without human annotators. 

To assess annotation quality, we compared inter-rater reliability across three sets: 400 pieces from the \citealp{davidson2017automated} dataset following general definitions, our descriptive annotations simulating \citealp{davidson2017automated} annotations, and our prescriptive annotations on the same 400 pieces. LLMs, prompted based on the prescriptive framework, were used in place of professional annotators to simulate limited human resources. The experiments demonstrate the effectiveness of smaller models fine-tuned on LLM-based prescriptive annotations for a 1942-piece set, comparing their performance to models fine-tuned on unused \citealp{davidson2017automated} annotations. Key contributions and findings are outlined below:
\paragraph{1.} This research proposes a prescriptive annotation benchmark to enable consistent offensive language data labeling with high reliability while preventing biases against language minorities, hence protecting natural language diversity. 
\paragraph{2.} This research contributes two newly annotated offensive language detection datasets created based on the proposed annotation benchmark \footnote{\href{https://github.com/Paparare/toxic_benchmark_2024}{Paparare/toxic\_benchmark\_2024}}.
\paragraph{3.} The proposed criteria lead to a higher inter-annotator agreement and reliability between prescriptive human annotations and between prescriptive human annotations and annotation generated by LLMs with prescriptive prompts derived from the annotation benchmark, compared to the original annotations based on vague and descriptive annotation instructions.
\paragraph{4.} Smaller models fine-tuned on a multi-source dataset annotated by LLMs outperform models trained on a single, significantly larger dataset annotated by humans, showing the effectiveness of structured guidelines in maintaining performance with limited data size and heterogeneous language types.

\section{Related Works}

\subsection{Common Annotation Bias in Past Datasets}
The issue of non-offensive language being mislabeled as offensive is also called unintended bias \cite{dixon2018measuring} or, more specifically, lexical bias \cite{garg2023handling} or linguistic bias \cite{mediabias2019} \cite{Tan2019AssessingSA}. For example, both (1) and (2) were identified as offensive:
\begin{quote}
(1) And apparently I'm committed to going to a new level since I used the key. Well FUCK. Curiosity killed the Cat(hy) \cite{barbieri2020tweeteval}
\end{quote}
\begin{quote}
(2) I ain't never seen a bitch so obsessed with they nigga\&\#128514;" I'm obsessed with mine \&\#128529 \cite{davidson2017automated}
\end{quote}
In (1), F**K is used as emotional emphasis. Similarly, slang does not always induce toxicity, as presented in (2); race-related term n***a is a neutral word often found in African American English (AAE) and gender-related b***h. The appropriateness of these terms varies, and their potential to harm others depends on their perlocutionary effect, influenced by the context and circumstances of use and reception \cite{allan2015slur, rahman2012n}.

\subsection{Annotation Paradigms}
Contextual swearing and minority language pose major challenges to simplistic judgments relying solely on phrasal units and general definitions \cite{pamungkas2023investigating, deas2023evaluation}. Simple reminders of exceptions and rare cases are insufficient, as unrestricted context interpretation based on individual assumptions inevitably introduces biases \cite{Rast2009ContextAI}. Educative annotation decisions regarding context must follow predefined instructions \cite{Giunchiglia2017PersonalCM, rottger2021two}. Descriptive data annotation embraces subjectivity to gain insights into diverse viewpoints but faces challenges in effectively eliciting, representing, and modeling those viewpoints \cite{rottger2021two, Alexeeva2023AnnotatingAT}. Prescriptive data annotation standardizes annotated features to provide consistent views of targeted language usages but risks overlooking some acceptable interpretations \cite{rottger2021two, Ruggeri2023OnTD}. Mitigating the potential deficiency of prescriptive annotation paradigms is a major issue in establishing this new benchmark.

\subsection{Studies-Driven Definition for Toxic Language}
Toxic language, a broader term than hate speech, refers to harm-inflicting expressions \cite{Buell1998ToxicD, Radfar2020CharacterizingVI, Baheti2021JustSN}. Hate speech, characterized by emotional and direct aggression towards targets \cite{Gelber2019TerroristExtremistSA, Elsherief2018HateLA}, is a manifestation of toxic language rather than being equivalent to it \cite{fortuna2020toxic}. Treating toxicity and hatred separately avoids potential confusion arising from treating them as interchangeable concepts. Offensiveness and toxicity in language are characterized by their capacity to evoke negative reactions, distinct from mere swear word usage \cite{legroski2018offensiveness}, and are tied to linguistic politeness and social decorum \cite{archard2014insults}, emphasizing the intention to denigrate rather than actual harm inflicted \cite{archard2008disgust}. Aggressiveness, while fundamental to dominating behavior \cite{Kacelnik1998PrimacyOO}, differs from outward toxicity that adversely impacts others. Aggressive components may contribute to offensive speech only when coupled with explicit intents to cause harm or distress \cite{stokes1970aggressive}. In short, toxic offensive language is the language that shows explicit aggression towards others. Separating language aggression from language intent aims to direct human judgment in annotation onto relevant textual features, avoiding biases and improving agreement by not erroneously marking provocative but ultimately inoffensive speech as inappropriate.

\section{Methodology}

\begin{table*}
\centering
\resizebox{\textwidth}{!}{%
\begin{tabular}{llll}
\hline
\textbf{Level} & \textbf{Item} & \textbf{Category} & \textbf{Example}\\
\hline
Lexical & Aggressive Noun Phrase and & \textit{Aggressive Item} & Stereotyped noun phrase/determiner phrase \\
 & Determiner Phrase & & (nigga, chingchong, \textit{etc.}), bitch, shit, dumbass, \textit{etc.}\\[.2\normalbaselineskip]
Lexical & Aggressive Verb Phrase & \textit{Aggressive Item }& fuck, hate, \textit{etc.} \\[.2\normalbaselineskip]
Lexical & Aggressive Adjective Phrase & \textit{Aggressive Item} & retarded, psycho, stupid, \textit{etc.}\\[.2\normalbaselineskip]
Lexical & Aggressive Adverb Phrase & \hl{\textit{Aggression Catalyze}r} & fucking, \textit{etc.}\\[.2\normalbaselineskip]
Syntactic & Strong Expression & \hl{\textit{Aggression Catalyzer}} & should, must, definitely, \textit{etc.} \\[.2\normalbaselineskip]
Syntactic & Rhetorical Question & \hl{\textit{Aggression Catalyzer}} & Doesn't everyone feel the same? \textit{etc.} \\[.2\normalbaselineskip]
Syntactic & Imperative & \hl{\textit{Aggression Catalyzer}} & Shut the door, \textit{etc.}\\[.5\normalbaselineskip]
Discourse & Ironic Expression & \hl{\textit{Aggression Catalyzer}} & Clear as mud, \textit{etc.}\\[.5\normalbaselineskip]
Discourse & False Construct & \textit{Aggressive Item} or & Those are people who only believe in \\[.2\normalbaselineskip]
& & \hl{\textit{Aggression Catalyzer}} & flat earth, \textit{etc.}\\[.2\normalbaselineskip]
Discourse & Controversial Content & \textit{Aggressive Item} & Inappropriate Content (adult, religious, \\
 & & & \textit{etc.}), jeering at others' mistakes \\
 & & & or misfortunes, \textit{etc.}\\
\hline
\end{tabular}
}
\caption{\label{aggression}
\textbf{Relative Aggression Scoring Reference}: Assigns numerical values for aggressive speech: 1 point for Aggressive Items (overtly toxic statements) and 0.5 points for Aggression Catalyzers (toxicity booster). The false construct will be an exception.
}
\end{table*}

Two components need to be assessed to determine toxicity: the direction of language intent (DI) and the presence of aggression (AG). DI has two labels: 1 for explicitly targeting other people and 0 for other cases. AG has three labels: 0 for non-aggressive, 1 for mildly aggressive, and 2 for intensely aggressive. A piece of text is categorized as \textbf{toxic or offensive if and only if it is labeled as 1 for DI and either 1 or 2 for AG.} The logic form is shown as follows:
\begin{align*}
\forall x \, (\text{Toxic}(x) &\iff (\text{DI}(x) = 1) \land \\
&(\text{AG}(x) = 1 \lor \text{AG}(x) = 2))
\end{align*}

\subsection{Annotation Criteria}
\textbf{Direction of Intent (DI)} indicates whether the language is directed externally (label 1) or not (label 0). Text segments receive a label of 1 if they directly refer to or address a specific person or group using second-person pronouns, proper nouns, or clear contextual references that signal an interpersonal attack or criticism. Text segments receive a label of 0 if the statements implicate others more implicitly, as is common with ironic expressions, or focus primarily on the speaker themselves. This simplified dichotomization aims to delineate clear instances of directive aggressive speech from more ambiguous cases. Since a tweet may contain multiple sentences with shifting targets, keeping disagreement in annotations is necessary for overlooking possible interpretations. \\
\textbf{Aggression (AG)} is annotated by categorizing negative, rude, or hostile attitudes into three levels: non-aggression (label 0, score 0), mild aggression (label 1, score 1), and intense aggression (label 2, score interval (1, $\infty$)). Table \ref{aggression} provides a relative score reference for categorizing and quantifying linguistic aggression across lexical, syntactic, and discourse levels. Linguistic items are classified as aggressive items (AI) that independently convey aggression or aggression catalyzers (AC) that intensify aggression but are not inherently aggressive. AIs (e.g., slurs, vulgarities, inflammatory content) are weighted 1 point, and ACs (e.g., emphatic language, rhetorical questions, imperatives, ironic expressions) 0.5 points. False constructs, which lead to flawed evaluations or unfair treatment, become AIs when paired with ACs but are still worth 0.5 points. In calculating the relative aggression score, each unique linguistic item should be counted only once, as including multiple items from one category does not typically increase aggressiveness. Lastly, to reduce the risk of overlooking possibilities, we encouraged annotators to keep different interpretations of ACs, as they are usually more implicit and open to various interpretations.

\subsection{Case Study}

The following two case studies will demonstrate how our proposed annotation guidelines help mitigate biases by providing a clear framework for assessing the direction of intent (DI) and the level of aggression (AG).

Example (1) demonstrates casual language usage: "And apparently I'm committed to going to a new level since I used the key. Well, FUCK. Curiosity killed the Cat(hy)" \cite{barbieri2020tweeteval}. We apply our annotation criteria to assess its toxicity. This example includes the aggressive verb phrase F**K, categorized as an aggressive item (AI), leading to an aggression score of 1, which indicates mild aggression. However, since the statement does not explicitly target any individual, its DI (Directed Insult) is labeled as 0. According to our criteria, a text is considered toxic or offensive only if it has a DI label of 1 and an AG label of either 1 or 2. Thus, example (1) is classified as non-toxic.

Example (2) illustrates the use of non-mainstream African American English: "I ain't never seen a bitch so obsessed with they nigga\&\#128514. I'm obsessed with mine\&\#128529" \cite{davidson2017automated}. This example contains two aggressive noun phrases ("b***h" and "n***a"), both categorized as AI. However, according to our guidelines, each unique linguistic item is counted only once when calculating the aggression score, resulting in an aggression score of 1, indicating mild aggression. Additionally, as the statement does not explicitly target another individual, its DI is labeled as 0. Despite the use of aggressive language, the absence of explicit targeting results in a non-toxic classification based on our annotation criteria.

\subsection{Human Annotation}

Two separate annotation processes were conducted, one with predefined criteria and one without. For the non-criteria-based human annotation, two annotators were given the question prompt, "Is the tweet toxic or offensive? If toxic or offensive, label 1; if it is not, label 0."  allow unrestricted subjectivity 
, following the descriptive data annotation paradigm. To examine the reliability of the original annotation, two annotators with academic backgrounds were chosen to resemble the diverse and unspecified backgrounds of CrowdFlower(CF) workers who were randomly employed and coded for \citealp{davidson2017automated}. The first annotator was a graduate marketing student familiar with internet culture but with no formal linguistic knowledge. The second was a graduate linguistics student with sufficient linguistic knowledge and socio-linguistic practices. Choosing annotators this way allowed evaluation of the reliability between the original and the descriptive data annotation under similar annotation conditions. The annotation with criteria was conducted by two linguistics graduate students who were trained with prescriptive instructions as presented in Appendix \ref{code} 
. Please find more information about annotators and more details about the annotation process in Appendix \ref{anno}.

\subsection{LLM Annotation}
Leveraging in-context learning is a promising approach to mitigate various learning biases while ensuring low-cost and highly generalizable processing \cite{lampinen2022can, margatina2023active, codaforno2023metaincontext}. Few-shot learning enables language models to rapidly adapt to new downstream tasks by analyzing a small set of relevant examples or interactions to discern expected outputs without extensive retraining \cite{gao2020making, perez2021true, mahabadi2022perfect}.

This study uses GPT-3.5-turbo and GPT-4 to generate prototypical responses with proposed criteria prompts. GPT-3.5's extensive architecture allows it to grasp and generate contextually relevant responses with limited input \cite{Yang2021AnES}. GPT-4 further enhances this capability due to its even more extensive training and sophisticated design \cite{openai2023gpt4}. We accessed both models via APIs to use small amounts of task-specific instruction to adapt to this task. Unlabeled data were processed with carefully constructed prompts to generate annotations consistent with pre-established formats. For descriptive LLM annotation, the question prompt used for human annotation was directly entered. For criteria-based LLM annotation, prompts were designed separately for the direction of intent, aggression recognition, and aggression scoring. The direction of intent prompt used general prescriptive instructions, while the aggression level prompt combined prescriptive instructions with few-shot examples sourced from the 'AI' and 'AC' categories to demonstrate specific scenarios. Given the subjective nature of aggression, including some examples in the latter prompt was crucial for ensuring some uniformity in annotations. Additionally, the challenge of neurotoxic degeneration is tackled by employing a method similar to Instruction Augmentation (INST) \cite{Prabhumoye2023AddingID}. We divided the aggression level prompt into two sections: one for assessing language use and another for aggression scoring. This division adheres to INST principles, enhancing the clarity and precision of instructional prompts for saving effects in cleaning the outcomes. 

\section{Data Analysis}

We randomly collected 400 tweets from the Offensive and Hate Speech dataset of the Davidson 2017 dataset \cite{davidson2017automated}. This dataset contains a high frequency of various types of offensive language and non-mainstream English. We chose this dataset because its dense toxic content and casual language use make it relatively straightforward for both human annotators and language models to process. The prevalence of clear toxic content reduces potential confusion and ambiguity that could skew the analysis.

\subsection{Inter-Annotator Agreement and Validation Analysis}

\begin{table}
\centering
{\small
\begin{tabular}{lccc}
\hline
\textbf{Pair} & \textbf{CK} & \textbf{AC1} & \textbf{Agr.\%} \\
\hline
\textbf{\textit{Descriptive}} \\
1T \& 2T & 0.5172 &  0.5094 & 76.50 \\
\hline
\textbf{\textit{Prescriptive \& Descriptive}} \\
1T \& 1T\_C & 0.3000 &  0.2406 & 66.75 \\
2T \& 1T\_C &  0.3889 &  0.3718 & 75.75 \\
1T \& 2T\_C & 0.2883 &   0.2229 & 66.25 \\
2T \& 2T\_C &  0.3966 & 0.3769 & 76.25 \\
\hline
\textbf{\textit{Prescriptive}} \\
1AG\_C \& 2AG\_C & 0.8422 & 0.8419 & 90.75 \\
1DI\_C \& 2DI\_C & 0.5913 & 0.5908 &  91.50\\
1T\_C \& 2T\_C & 0.7487 & 0.7486 & 92.50 \\
\hline
\end{tabular}
}
\caption{\textbf{Inter-Annotator Reliability Evaluation for Prescriptive and Descriptive Annotations}:
1T denotes descriptive toxicity, marketing student;
2T denotes descriptive toxicity, linguistics student;
1AG\_C denotes prescriptive aggression, Annotator 1;
2AG\_C denotes prescriptive aggression, Annotator 2;
1DI\_C denotes prescriptive intent direction, Annotator 1;
2DI\_C denotes prescriptive intent direction, Annotator 2;
1T\_C denotes prescriptive toxicity, Annotator 1;
2T\_C denotes prescriptive toxicity, Annotator 2}
\label{tab:cohen_gwet_values}
\end{table}

\begin{table*}
\centering

\begin{tabular}{lccc}
\hline
\textbf{Pair} & \textbf{CK} & \textbf{AC1} & \textbf{Agr. \%} \\
\hline

1T \& \citealp{davidson2017automated} & -0.0475 & -0.2552 & 51.25 \\
2T \& \citealp{davidson2017automated} & -0.0566 & -0.1742 & 62.25 \\
\hline \hline
1T\_C \& \citealp{davidson2017automated} & -0.0884 & -0.1237 & 75.00 \\

2T\_C \& \citealp{davidson2017automated} & -0.0405 & -0.0698 & 77.00 \\
\hline
\end{tabular}
\caption{Inter-annotator Reliability Evaluation on prescriptive, descriptive, and original annotation.}
\label{tab:original}
\end{table*}

Confusion matrices for all annotations are listed in Appendix \ref{sec:appendix}, and the distributions are displayed in Appendix \ref{sec:appendix2}. For a comprehensive evaluation of annotator consistency, we calculated Cohen's Kappa (CK) \cite{mchugh2012interrater} and Gwet's AC1 (AC1)\cite{cicchetti1976assessing}, as detailed in Table \ref{tab:cohen_gwet_values}. Initially, we assessed the inter-annotator reliability for both our annotations without criteria and those from \citealp{davidson2017automated}, displayed in Table \ref{tab:original}. Gwet's AC1 can help avoid the paradoxical behavior and biased estimates associated with Cohen's Kappa, especially in situations of high agreement and prevalence \cite{zec2017suppl}. 

According to Table \ref{tab:cohen_gwet_values}, incorporating specific criteria in the annotation process significantly enhances consistency and agreement between raters. This conclusion is supported by the larger positive values of trinary metrics for with-criteria pairs compared to without-criteria pairs and with-without-criteria pairs. Cohen's Kappa and Gwet's AC1 values, which adjust for chance agreement, indicate only moderate agreement without criteria. However, these values markedly increased when criteria were applied, as the first and last pairs approached near-perfect agreement levels. This underscores the critical role of well-defined criteria in enhancing reliability and validity of qualitative assessments. Interestingly, the reliability evaluations for with-without-criteria pairs are even lower than without-criteria pairs, suggesting the annotation logic for the two annotation types are entirely different. 

Unlike our annotations, the comparison with the original annotations presents contrasting results in Table \ref{tab:original}. Cohen's Kappa and Gwet's AC1 values are negative across all comparisons, suggesting a level of disagreement more pronounced than random chance. This also indicates underlying distinctions in how the annotations were carried out, and the fact that the majority vote labels they used for the final label were not from the same annotator could be a reason why reliability tests exhibit so much difference. These statistics starkly contrast the earlier findings where criteria application resulted in a near-perfect agreement for specific pairs. Although the agreement percentages showed some surface agreement, they do not align with the deeper discordance indicated by the negative Cohen's Kappa and Gwet's AC1 values. As a result, prescriptive data annotations (1T\_C, 2T\_C) show higher reliability compared to descriptive data annotations (1T, 2T). Prescriptive data annotation paradigms are more appropriate for this task. This discrepancy highlights the complexities in achieving inter-rater reliability and the need to thoroughly review annotation guidelines and processes to understand and rectify the significant misalignments.

\begin{table*}
\centering
\resizebox{0.9\textwidth}{!}{%
\begin{tabular}{lccclccc}
\hline
\textbf{Pair} & \textbf{CK} & \textbf{AC1} & \textbf{Agr. \%} & \textbf{Pair} & \textbf{CK} & \textbf{AC1}& \textbf{Agr. \%}\\
\hline
\multicolumn{8}{l}{\textbf{\textit{Without Criteria}}} \\
1T \& G4T & 0.2030 & 0.0685 & 62.75  &  1T \& G3T & 0.3149 & 0.2532   &  \textbf{67.50}\\
2T \& G4T & 0.2819 & 0.2190 & 73.75  &   2T \& G3T &  0.3534 &  0.3331 &  \textbf{74.50} \\
\hline \hline
\multicolumn{8}{l}{\textbf{\textit{With Criteria}}} \\
1DI\_C \& G4DI\_C & 0.3376 & 0.3361 &  87.00 & 1DI\_C \& G3DI\_C &  0.1999 & 0.1799 & \textbf{87.75}\\
2DI\_C \& G4DI\_C & 0.5647 & 0.5646 & \textbf{92.25} & 2DI\_C \& G3DI\_C & 0.2820 & 0.2704 & 90.25\\
1AG\_C \& G4AG\_C & 0.3460 &  0.3016 & \textbf{62.5} & 1AG\_C \& G3AG\_C & 0.2813 & 0.2605 & 59.25\\
2AG\_C \& G4AG\_C & 0.3849 & 0.3565  & \textbf{66.5} & 2AG\_C \& G3AG\_C & 0.2700 & 0.2588 & 60.0 \\
1T\_C \& G4T\_C & 0.5299 & 0.5282 & \textbf{87.00}  & 1T\_C \& G3T\_C & 0.4013 & 0.3887 & 85.5\\
2T\_C \& G4T\_C & 0.6103 & 0.6094 & \textbf{89.50} & 2T\_C \& G3T\_C & 0.4015 & 0.3910 & 86.0 \\
\hline
\end{tabular}
} 
\caption{\textbf{Inter-Annotator Reliability Evaluation of GPT Annotations and Human Annotations}: G4T denotes descriptive toxicity, GPT-4; G3T denotes descriptive toxicity, GPT-3.5-turbo; G4DI\_C denotes prescriptive intent direction, GPT-4; G4AG\_C denotes prescriptive aggression, GPT-4; G4T\_C denotes prescriptive toxicity, GPT-4; G3DI\_C denotes prescriptive intent direction, GPT-3-turbo; G3AG\_C denotes prescriptive aggression, GPT-3.5-turbo; G3T\_C denotes prescriptive toxicity, GPT-3.5-turbo}
\label{tab:agreement_info}
\end{table*}

\begin{table*}[h]
\centering
\begin{tabular}{lccc}
\hline
\textbf{Model (Fine-Tuning Data)} & \textbf{DI (F1)} & \textbf{AG (F1)} & \textbf{T (F1)} \\ \hline
RoBERTa-base (\citealp{davidson2017automated}) & - & - & 0.912 \\ 
DeBERTa-base (\citealp{davidson2017automated}) & - & - &  0.908\\ 
\hline
RoBERTa-base (G3P) &  0.894 & 0.656 & - \\ 
DeBERTa-base (G3P) & 0.913 &  0.715 & - \\ 
\hline
RoBERTa-base (G4P) & 0.927 & 0.849 & - \\ 
DeBERTa-base (G4P) & 0.925 & 0.825 & - \\ 
\hline
\end{tabular}
\caption{Learning Performance for BERT models Fine-tuned on \citealp{davidson2017automated} baseline and GPT-annotated Datasets with Macro-averaged F1}
\label{acc}
\end{table*}

\begin{table*}[h]
\centering
\resizebox{0.94\textwidth}{!}{%
\begin{tabular}{lccccccc}
\hline
\textbf{Model (Fine-Tuning Data)} &  &  &  &  &  \textbf{1T} &  \textbf{2T} \\ \hline
RoBERTa-base (\citealp{davidson2017automated}) &  &  &  &  & 0.379 & 0.665  \\ 
DeBERTa-base (\citealp{davidson2017automated}) &  &  &  &  & 0.379 & 0.531\\ \hline \hline
 & \textbf{1DI\_C} & \textbf{2DI\_C} & \textbf{1AG\_C} & \textbf{2AG\_C} &  \textbf{1T\_C} &  \textbf{2T\_C} \\ \hline
RoBERTa-base (\citealp{davidson2017automated}) & - & - & - & - & 0.728 & 0.742\\ 
DeBERTa-base (\citealp{davidson2017automated}) & - & - & - & - & 0.728 & 0.742 \\ 
\hline 
RoBERTa-base (G3P) & 0.828 & 0.867 & \textbf{0.597} & \textbf{0.572} & 0.806 & 0.819 \\ 
DeBERTa-base (G3P) & 0.839 & 0.877 & 0.525 & 0.558 & 0.793 & 0.811 \\
\hline 
RoBERTa-base (G4P) & 0.850 & 0.889 & 0.389 & 0.446 & \textbf{0.837} & \textbf{0.859}  \\ 
DeBERTa-base (G4P) & \textbf{0.879} & \textbf{0.908} & 0.383 & 0.441 & 0.817 & 0.839 \\
\hline
\end{tabular}}
\caption{Macro-averaged F1 Scores of BERT models fine-tuned on \citealp{davidson2017automated} baseline and GPT-annotated data in Comparison with Human Annotations}
\label{performance}
\end{table*}

\subsection{Validation and Agreement Analysis of Human and GPT Annotations}

As Cohen's Kappa and Gwet's AC1 were created to assess inter-rater reliability between human annotators, directly applying them to evaluate agreement between machine and human annotations may not be entirely apt \cite{popovic2021reproduction}. While primarily intended for only human judgment scenarios, we include evaluations using these metrics when comparing GPT model predictions and human labels since dedicated methods for assessing machine-human agreement have yet to be established. We analyzed the concordance between human annotations and those generated by GPT 
models, namely GPT-4 \cite{openai2023gpt4} and GPT-3.5 \cite{gpt3.5}, across two annotation categories. 
The trinary evaluations in Table \ref{tab:agreement_info} demonstrate reasonable consistency and agreement between human annotations and those from GPT-3.5 and GPT-4. Without prompted criteria, GPT-3.5 slightly outperforms GPT-4 in both agreement and reliability, but refining the prompts enabled more effective and reliable synergy between automated toxicity analysis and human-like interpretation. Using the proposed criteria significantly improved the alignment with human judgment for both models, especially for GPT-4 annotations.  Inter-rater reliability Under criteria-based scenarios, GPT-4 annotations showed comparable agreement and consistent inter-rater reliability. The reliability statistics show that GPT annotations have even higher agreement and consistency than the original human annotations and without-criteria human annotations following the descriptive paradigm. The established criteria improved accuracy. Additionally, GPT-4 outperformed GPT-3.5 on this task. This suggests an aptitude for criteria-based analysis. After implementing the proposed criteria, these notable improvements demonstrate that prescriptive data annotation instructions can help researchers overcome the lack of human annotator resources.

\section{Experiments}

The experiment settings involve fine-tuning two models, RoBERTa-base with approximately 125 million parameters \cite{DBLP:journals/corr/abs-1907-11692} and DeBERTa-base with approximately 139  million parameters \cite{he2021deberta}, using a training batch size of 8 and an evaluation batch size of 16 with 5e-5 learning rate. The models are trained for 3 epochs, with the dataset split into 90\% for training and 10\% for testing. To stabilize training, a learning rate warmup strategy is employed with 500 warmup steps. Weight decay regularization with a value of 0.01 is applied to prevent overfitting by encouraging smaller weights. Two datasets were used in this study. The baseline models were fine-tuned on 2,438 tweets from the Davidson 2017 dataset \cite{davidson2017automated}, excluding 400 pieces used in statistical analysis. In comparison, a 1,942-piece dataset was compiled for prescriptive LLM annotations, consisting of 295 Reddit posts in African American English \cite{deas2023evaluation}, 341 tweets from OLID \cite{zampieri2019predicting}, 311 tweets from the offensive and hate speech dataset \cite{davidson2017automated}, and 1,000 tweets from Hateval \cite{basile2019semeval}. The combination of different datasets helps mitigate extrusive language features, while the inclusion of diverse social media platforms (e.g., Reddit, Twitter) facilitates robust exposure to various language types and dialects. Previous studies and empirical observations suggest that larger datasets, particularly those with language types similar to the target application, tend to lead to higher performance in language models \cite{sahlgren2016effects, linjordet2019impact, kaplan2020scaling}. Therefore, the Davidson 2017 dataset, with its size and domain relevance advantages, would likely enable superior performance compared to the smaller, more complex 1,942-piece dataset.

\subsection{Result Analysis and Discussion}

As shown in Table \ref{acc}, when fine-tuned on different datasets, DeBERTa-base slightly outperforms RoBERTa-base on the baseline dataset, achieving macro F1 scores of 0.908 and 0.912, respectively. However, RoBERTa-base achieves higher accuracy in prescriptive Aggression (AG) and prescriptive Direction of Intent (DI) when trained on GPT-annotated datasets (G3P\footnote{1,942-piece set annotated by GPT-3.5-turbo with proposed criteria} and G4P\footnote{1,942-piece set annotated by GPT-4 with proposed criteria}). RoBERTa-base achieves macro F1 scores of 0.894 and 0.656 for DI and AG, respectively, on the G3P dataset and 0.927 and 0.849 on the G4P dataset. All experiments were conducted using an NVIDIA A100 GPU. Macro-F1 scores in Table \ref{performance} indicate that fine-tuned models align well with human annotations in identifying language intent (1DI\_C and 2DI\_C) but struggle more with aggression classifications (1AG\_C and 2AG\_C). When fine-tuned on the baseline dataset, BERT models moderately agree with human toxicity annotations (1T and 2T), with macro F1 scores of 0.379 for 1T and 0.665 and 0.531 for 2T using RoBERTa-base and DeBERTa-base, respectively. Notably, criteria-based auto-annotations improve model performance, with higher agreement rates using the G4P dataset. Models fine-tuned on G4P annotations achieved lower macro F1 scores for aggression (0.389 and 0.446 for 1AG\_C and 2AG\_C using RoBERTa-base) but higher macro F1 scores for toxicity (0.837 and 0.859 for 1T\_C and 2T\_C using RoBERTa-base).

These results suggest that GPT-4's annotations may not have captured the features needed to distinguish between mild and intense aggression. Still, they did exhibit features that differentiate non-aggressive from aggressive content. The similar and higher macro F1 scores for toxicity in models fine-tuned on G3P and G4P (ranging from 0.793 to 0.859) compared to baselines demonstrate the effectiveness of using properly-prompted LLMs over random human annotators. Despite improvements, fine-tuned BERT models still lag behind prescriptive human annotators and prescriptively-prompted LLM annotations, possibly due to small dataset sizes. This result contradicts the previous hypothesis that the baseline dataset with a much larger size and more uniform language patterns would help small models outperform LLM annotations; instead, it strongly suggests the robustness of models fine-tuned on prescriptively annotated data.

\section{Conclusion}
In conclusion, this study improves offensive language detection by introducing a prescriptive annotation benchmark that separately evaluates intent and aggression, reducing bias and preserving language diversity. Our analysis demonstrates that LLMs, guided by few-shot learning and clear criteria, can identify annotation errors in casual and non-mainstream language, offering better reliability than previous studies. The proposed framework also improves BERT's performance on small, complex datasets, outperforming baselines in resource-limited scenarios. These findings highlight the efficiency of this approach in optimizing data use and adapting toxic content moderation systems to diverse language patterns, even with limited annotation resources.

\section*{Limitations}

First of all, aggressive expression classifications are not definitive. There is room for different interpretations to mitigate the risk of over-generalization associated with prescriptive annotation. What constitutes a specific category of aggression could shift over time as cultural norms and language use evolve. Additionally, it can sometimes be difficult to precisely categorize certain expressions of aggression due to variations in language, influences from popular culture, and other contextual factors. The following criteria only try to grasp a more objective overview of aggression, which does not intend to rule out all subjectivity.  Putting values on categories assesses the functional diversity of different language components, providing a more precise evaluation of the aggression level. However, in certain instances, merely adding more terms from a single category can decrease the perceived aggression. This is because excessive repetition of similar aggressive language might come across as impotent rage, reducing the overall impact of the aggression expressed.

We identified some limitations that are important for guiding future research. While prescriptive annotation paradigms may better identify uniform patterns, they risk overlooking meaningful interpretations not yet recognized by linguists and social scientists. The proposed criteria account for variations in English, but their practical application relies heavily on annotators' language knowledge. The dynamic nature of internet language poses additional challenges for human coders to accurately comprehend tweets, as no annotators can fully grasp the breadth of English online language, let alone code-switching usages by multilingual users. On the other hand, annotators lacking contextual understanding of in-group language may erroneously analyze utterances meant to promote within-community comprehensibility, a limitation challenging to resolve through improved annotation design. In contrast, LLMs demonstrate an advantage in aggregating insights from considerably larger data sources. Therefore, determining approaches for incorporating LLMs in detection alongside human rationale remains an important direction for further research. 

Furthermore, the scope of human annotation within our dataset could be expanded. Human annotation of a dense toxicity corpus reveals high agreement; however, corpora containing more implicit cultural-related expressions would likely yield lower agreement rates. So, the human agreement in this research is only a reference, not a solid upper bound. Although we relied on a significant amount of human input, the complexities and nuances of offensive language suggest that a broader and more diverse set of human annotations could enhance the model's understanding and accuracy. Another limitation lies in the size of our auto-annotated dataset. Additionally, there is room for improvement in the performance of smaller models on the automatically generated dataset. Open-source LLMs could be possible substitutes. Exploring different configurations, experimenting with various model architectures, and further tuning could enhance performance. 


\section*{Acknowledgement}
I would like to sincerely thank Hao Yu and Kedi Mo from Teachers College, Columbia University, and Jiaqi Wang from the Olin Business School of Washington University in St. Louis for their invaluable help with the annotations for this project. Their hard work and dedication were crucial to its completion. Special thanks are due to Nicholas Deas and Professor Kathleen McKeown from the Fu Foundation School of Engineering and Applied Science at Columbia University, who guided me in conducting research and designing the research framework. I would also like to express my gratitude to Dr. Howard A. Williams and Dr. Payment Vafaee from Teachers College, Columbia University, whose insightful brainstorming sessions and valuable discussions helped shape the boundary between aggressive language and toxic language, providing further insight into addressing toxic language. Their guidance was instrumental in refining the framework for this study. I would also like to extend my sincere thanks to Dr. Erik Voss from Teachers College, Columbia University, my MA project advisor, for his unwavering support and guidance throughout the drafting of this research. Finally, I would like to thank the reviewers for their thoughtful suggestions, which helped improve the clarity of the paper.

\bibliography{anthology,custom}

\begin{thebibliography}{58}
\expandafter\ifx\csname natexlab\endcsname\relax\def\natexlab#1{#1}\fi

\bibitem[{Alexeeva et~al.(2023)Alexeeva, Hyland, Alcock, Cohen, Kanyamahanga, Anni, and Surdeanu}]{Alexeeva2023AnnotatingAT}
Maria Alexeeva, Caroline Hyland, Keith Alcock, Allegra Argent~Beal Cohen, Hubert Kanyamahanga, Isaac~Kobby Anni, and Mihai Surdeanu. 2023.
\newblock \href {https://api.semanticscholar.org/CorpusID:260063070} {Annotating and training for population subjective views}.
\newblock In \emph{Workshop on Computational Approaches to Subjectivity, Sentiment and Social Media Analysis}.

\bibitem[{Allan(2015)}]{allan2015slur}
Keith Allan. 2015.
\newblock When is a slur not a slur? the use of nigger in ‘pulp fiction’.
\newblock \emph{Language Sciences}, 52:187--199.

\bibitem[{Archard(2008)}]{archard2008disgust}
David Archard. 2008.
\newblock Disgust, offensiveness and the law.
\newblock \emph{Journal of Applied Philosophy}, 25(4):314--321.

\bibitem[{Archard(2014)}]{archard2014insults}
David Archard. 2014.
\newblock Insults, free speech and offensiveness.
\newblock \emph{Journal of Applied Philosophy}, 31(2):127--141.

\bibitem[{Baheti et~al.(2021)Baheti, Sap, Ritter, and Riedl}]{Baheti2021JustSN}
Ashutosh Baheti, Maarten Sap, Alan Ritter, and Mark~O. Riedl. 2021.
\newblock \href {https://api.semanticscholar.org/CorpusID:237303836} {Just say no: Analyzing the stance of neural dialogue generation in offensive contexts}.
\newblock \emph{ArXiv}, abs/2108.11830.

\bibitem[{Barbieri et~al.(2020)Barbieri, Camacho-Collados, Espinosa-Anke, and Neves}]{barbieri2020tweeteval}
Francesco Barbieri, Jose Camacho-Collados, Luis Espinosa-Anke, and Leonardo Neves. 2020.
\newblock {TweetEval:Unified Benchmark and Comparative Evaluation for Tweet Classification}.
\newblock In \emph{Proceedings of Findings of EMNLP}.

\bibitem[{Basile et~al.(2019)Basile, Bosco, Fersini, Nozza, Patti, Pardo, Rosso, and Sanguinetti}]{basile2019semeval}
Valerio Basile, Cristina Bosco, Elisabetta Fersini, Debora Nozza, Viviana Patti, Francisco Manuel~Rangel Pardo, Paolo Rosso, and Manuela Sanguinetti. 2019.
\newblock Semeval-2019 task 5: Multilingual detection of hate speech against immigrants and women in twitter.
\newblock In \emph{Proceedings of the 13th international workshop on semantic evaluation}, pages 54--63.

\bibitem[{Buell(1998)}]{Buell1998ToxicD}
Lawrence Buell. 1998.
\newblock \href {https://api.semanticscholar.org/CorpusID:2619056} {Toxic discourse}.
\newblock \emph{Critical Inquiry}, 24:639 -- 665.

\bibitem[{Cicchetti(1976)}]{cicchetti1976assessing}
Domenic~V Cicchetti. 1976.
\newblock Assessing inter-rater reliability for rating scales: resolving some basic issues.
\newblock \emph{The British Journal of Psychiatry}, 129(5):452--456.

\bibitem[{Coda-Forno et~al.(2023)Coda-Forno, Binz, Akata, Botvinick, Wang, and Schulz}]{codaforno2023metaincontext}
Julian Coda-Forno, Marcel Binz, Zeynep Akata, Matthew Botvinick, Jane~X. Wang, and Eric Schulz. 2023.
\newblock \href {http://arxiv.org/abs/2305.12907} {Meta-in-context learning in large language models}.

\bibitem[{Davani et~al.(2023)Davani, Atari, Kennedy, and Dehghani}]{10.1162/tacl_a_00550}
Aida~Mostafazadeh Davani, Mohammad Atari, Brendan Kennedy, and Morteza Dehghani. 2023.
\newblock \href {https://doi.org/10.1162/tacl_a_00550} {{Hate Speech Classifiers Learn Normative Social Stereotypes}}.
\newblock \emph{Transactions of the Association for Computational Linguistics}, 11:300--319.

\bibitem[{Davani et~al.(2021)Davani, D'iaz, and Prabhakaran}]{Davani2021DealingWD}
Aida~Mostafazadeh Davani, M.~C. D'iaz, and Vinodkumar Prabhakaran. 2021.
\newblock \href {https://api.semanticscholar.org/CorpusID:238634750} {Dealing with disagreements: Looking beyond the majority vote in subjective annotations}.
\newblock \emph{Transactions of the Association for Computational Linguistics}, 10:92--110.

\bibitem[{Davidson et~al.(2017)Davidson, Warmsley, Macy, and Weber}]{davidson2017automated}
Thomas Davidson, Dana Warmsley, Michael Macy, and Ingmar Weber. 2017.
\newblock Automated hate speech detection and the problem of offensive language.
\newblock In \emph{Proceedings of the international AAAI conference on web and social media}, volume~11, pages 512--515.

\bibitem[{de~Gibert et~al.(2018)de~Gibert, Perez, Garc{\'\i}a-Pablos, and Cuadros}]{gibert2018hate}
Ona de~Gibert, Naiara Perez, Aitor Garc{\'\i}a-Pablos, and Montse Cuadros. 2018.
\newblock \href {https://doi.org/10.18653/v1/W18-5102} {{Hate Speech Dataset from a White Supremacy Forum}}.
\newblock In \emph{Proceedings of the 2nd Workshop on Abusive Language Online ({ALW}2)}, pages 11--20, Brussels, Belgium. Association for Computational Linguistics.

\bibitem[{Deas et~al.(2023)Deas, Grieser, Kleiner, Patton, Turcan, and McKeown}]{deas2023evaluation}
Nicholas Deas, Jessi Grieser, Shana Kleiner, Desmond Patton, Elsbeth Turcan, and Kathleen McKeown. 2023.
\newblock \href {http://arxiv.org/abs/2305.14291} {Evaluation of african american language bias in natural language generation}.

\bibitem[{Dixon et~al.(2018)Dixon, Li, Sorensen, Thain, and Vasserman}]{dixon2018measuring}
Lucas Dixon, John Li, Jeffrey Sorensen, Nithum Thain, and Lucy Vasserman. 2018.
\newblock Measuring and mitigating unintended bias in text classification.
\newblock In \emph{Proceedings of the 2018 AAAI/ACM Conference on AI, Ethics, and Society}, pages 67--73.

\bibitem[{Elsherief et~al.(2018)Elsherief, Kulkarni, Nguyen, Wang, and Belding-Royer}]{Elsherief2018HateLA}
Mai Elsherief, Vivek Kulkarni, Dana Nguyen, William~Yang Wang, and Elizabeth~M. Belding-Royer. 2018.
\newblock \href {https://api.semanticscholar.org/CorpusID:4809781} {Hate lingo: A target-based linguistic analysis of hate speech in social media}.
\newblock In \emph{International Conference on Web and Social Media}.

\bibitem[{Fan et~al.(2019)Fan, White, Sharma, Su, Choubey, Huang, and Wang}]{mediabias2019}
Lisa Fan, Marshall White, Eva Sharma, Ruisi Su, Prafulla~Kumar Choubey, Ruihong Huang, and Lu~Wang. 2019.
\newblock In plain sight: Media bias through the lens of factual reporting.
\newblock In \emph{Proceedings of the 2019 Conference on Empirical Methods in Natural Language Processing}. Association for Computational Linguistics.

\bibitem[{Fortuna et~al.(2020)Fortuna, Soler, and Wanner}]{fortuna2020toxic}
Paula Fortuna, Juan Soler, and Leo Wanner. 2020.
\newblock Toxic, hateful, offensive or abusive? what are we really classifying? an empirical analysis of hate speech datasets.
\newblock In \emph{Proceedings of the 12th language resources and evaluation conference}, pages 6786--6794.

\bibitem[{Founta et~al.(2018)Founta, Djouvas, Chatzakou, Leontiadis, Blackburn, Stringhini, Vakali, Sirivianos, and Kourtellis}]{founta2018large}
Antigoni Founta, Constantinos Djouvas, Despoina Chatzakou, Ilias Leontiadis, Jeremy Blackburn, Gianluca Stringhini, Athena Vakali, Michael Sirivianos, and Nicolas Kourtellis. 2018.
\newblock Large scale crowdsourcing and characterization of twitter abusive behavior.
\newblock In \emph{Proceedings of the international AAAI conference on web and social media}, volume~12.

\bibitem[{Gao et~al.(2020)Gao, Fisch, and Chen}]{gao2020making}
Tianyu Gao, Adam Fisch, and Danqi Chen. 2020.
\newblock Making pre-trained language models better few-shot learners.
\newblock \emph{arXiv preprint arXiv:2012.15723}.

\bibitem[{Garg et~al.(2023)Garg, Masud, Suresh, and Chakraborty}]{garg2023handling}
Tanmay Garg, Sarah Masud, Tharun Suresh, and Tanmoy Chakraborty. 2023.
\newblock \href {http://arxiv.org/abs/2202.00126} {Handling bias in toxic speech detection: A survey}.

\bibitem[{Gelber(2019)}]{Gelber2019TerroristExtremistSA}
Katharine Gelber. 2019.
\newblock \href {https://api.semanticscholar.org/CorpusID:195330839} {Terrorist-extremist speech and hate speech: Understanding the similarities and differences}.
\newblock \emph{Ethical Theory and Moral Practice}, pages 1--16.

\bibitem[{Geva et~al.(2019)Geva, Goldberg, and Berant}]{Geva2019AreWM}
Mor Geva, Yoav Goldberg, and Jonathan Berant. 2019.
\newblock \href {https://api.semanticscholar.org/CorpusID:201124736} {Are we modeling the task or the annotator? an investigation of annotator bias in natural language understanding datasets}.
\newblock \emph{ArXiv}, abs/1908.07898.

\bibitem[{Giunchiglia et~al.(2017)Giunchiglia, Bignotti, and Zeni}]{Giunchiglia2017PersonalCM}
Fausto Giunchiglia, Enrico Bignotti, and Mattia Zeni. 2017.
\newblock \href {https://api.semanticscholar.org/CorpusID:28900644} {Personal context modelling and annotation}.
\newblock \emph{2017 IEEE International Conference on Pervasive Computing and Communications Workshops (PerCom Workshops)}, pages 117--122.

\bibitem[{He et~al.(2021)He, Liu, Gao, and Chen}]{he2021deberta}
Pengcheng He, Xiaodong Liu, Jianfeng Gao, and Weizhu Chen. 2021.
\newblock \href {https://openreview.net/forum?id=XPZIaotutsD} {Deberta: Decoding-enhanced bert with disentangled attention}.
\newblock In \emph{International Conference on Learning Representations}.

\bibitem[{Kacelnik and Norris(1998)}]{Kacelnik1998PrimacyOO}
Alejandro Kacelnik and Sasha Norris. 1998.
\newblock \href {https://api.semanticscholar.org/CorpusID:9462611} {Primacy of organising effects of testosterone}.
\newblock \emph{Behavioral and Brain Sciences}, 21:365 -- 365.

\bibitem[{Kaplan et~al.(2020)Kaplan, McCandlish, Henighan, Brown, Chess, Child, Gray, Radford, Wu, and Amodei}]{kaplan2020scaling}
Jared Kaplan, Sam McCandlish, Tom Henighan, Tom~B. Brown, Benjamin Chess, Rewon Child, Scott Gray, Alec Radford, Jeffrey Wu, and Dario Amodei. 2020.
\newblock \href {http://arxiv.org/abs/2001.08361} {Scaling laws for neural language models}.

\bibitem[{Kuwatly et~al.(2020)Kuwatly, Wich, and Groh}]{Kuwatly2020IdentifyingAM}
Hala~Al Kuwatly, Maximilian Wich, and Georg Groh. 2020.
\newblock \href {https://api.semanticscholar.org/CorpusID:226283471} {Identifying and measuring annotator bias based on annotators’ demographic characteristics}.
\newblock In \emph{Workshop on Abusive Language Online}.

\bibitem[{Lampinen et~al.(2022)Lampinen, Dasgupta, Chan, Matthewson, Tessler, Creswell, McClelland, Wang, and Hill}]{lampinen2022can}
Andrew~K Lampinen, Ishita Dasgupta, Stephanie~CY Chan, Kory Matthewson, Michael~Henry Tessler, Antonia Creswell, James~L McClelland, Jane~X Wang, and Felix Hill. 2022.
\newblock Can language models learn from explanations in context?
\newblock \emph{arXiv preprint arXiv:2204.02329}.

\bibitem[{Legroski(2018)}]{legroski2018offensiveness}
Marina~Chiara Legroski. 2018.
\newblock Offensiveness scale: how offensive is this expression?
\newblock \emph{Estudos Lingu{\'\i}sticos (S{\~a}o Paulo. 1978)}, 47(1):169--180.

\bibitem[{Linjordet and Balog(2019)}]{linjordet2019impact}
Trond Linjordet and Krisztian Balog. 2019.
\newblock Impact of training dataset size on neural answer selection models.
\newblock In \emph{Advances in Information Retrieval: 41st European Conference on IR Research, ECIR 2019, Cologne, Germany, April 14--18, 2019, Proceedings, Part I 41}, pages 828--835. Springer.

\bibitem[{Liu et~al.(2019)Liu, Ott, Goyal, Du, Joshi, Chen, Levy, Lewis, Zettlemoyer, and Stoyanov}]{DBLP:journals/corr/abs-1907-11692}
Yinhan Liu, Myle Ott, Naman Goyal, Jingfei Du, Mandar Joshi, Danqi Chen, Omer Levy, Mike Lewis, Luke Zettlemoyer, and Veselin Stoyanov. 2019.
\newblock \href {http://arxiv.org/abs/1907.11692} {Roberta: {A} robustly optimized {BERT} pretraining approach}.
\newblock \emph{CoRR}, abs/1907.11692.

\bibitem[{Liu et~al.(2022)Liu, Richardson, Hatcher, and Prudhommeaux}]{Liu2022NotAA}
Zoey Liu, Crystal Richardson, Richard~J. Hatcher, and Emily Prudhommeaux. 2022.
\newblock \href {https://api.semanticscholar.org/CorpusID:248118721} {Not always about you: Prioritizing community needs when developing endangered language technology}.
\newblock In \emph{Annual Meeting of the Association for Computational Linguistics}.

\bibitem[{Mahabadi et~al.(2022)Mahabadi, Zettlemoyer, Henderson, Saeidi, Mathias, Stoyanov, and Yazdani}]{mahabadi2022perfect}
Rabeeh~Karimi Mahabadi, Luke Zettlemoyer, James Henderson, Marzieh Saeidi, Lambert Mathias, Veselin Stoyanov, and Majid Yazdani. 2022.
\newblock Perfect: Prompt-free and efficient few-shot learning with language models.
\newblock \emph{arXiv preprint arXiv:2204.01172}.

\bibitem[{Margatina et~al.(2023)Margatina, Schick, Aletras, and Dwivedi-Yu}]{margatina2023active}
Katerina Margatina, Timo Schick, Nikolaos Aletras, and Jane Dwivedi-Yu. 2023.
\newblock Active learning principles for in-context learning with large language models.
\newblock \emph{arXiv preprint arXiv:2305.14264}.

\bibitem[{McHugh(2012)}]{mchugh2012interrater}
Mary~L McHugh. 2012.
\newblock Interrater reliability: the kappa statistic.
\newblock \emph{Biochemia medica}, 22(3):276--282.

\bibitem[{Mollas et~al.(2020)Mollas, Chrysopoulou, Karlos, and Tsoumakas}]{mollas2020ethos}
Ioannis Mollas, Zoe Chrysopoulou, Stamatis Karlos, and Grigorios Tsoumakas. 2020.
\newblock Ethos: an online hate speech detection dataset.
\newblock \emph{arXiv preprint arXiv:2006.08328}.

\bibitem[{OpenAI(2022)}]{gpt3.5}
OpenAI. 2022.
\newblock Gpt-3.5: Language models are few-shot learners.
\newblock \url{https://openai.com/blog/gpt-3-5-update/}.
\newblock Accessed: [Insert current date here].

\bibitem[{OpenAI(2023)}]{openai2023gpt4}
OpenAI. 2023.
\newblock \href {http://arxiv.org/abs/2303.08774} {Gpt-4 technical report}.

\bibitem[{Pamungkas et~al.(2023)Pamungkas, Basile, and Patti}]{pamungkas2023investigating}
Endang~Wahyu Pamungkas, Valerio Basile, and Viviana Patti. 2023.
\newblock Investigating the role of swear words in abusive language detection tasks.
\newblock \emph{Language Resources and Evaluation}, 57(1):155--188.

\bibitem[{Perez et~al.(2021)Perez, Kiela, and Cho}]{perez2021true}
Ethan Perez, Douwe Kiela, and Kyunghyun Cho. 2021.
\newblock True few-shot learning with language models.
\newblock \emph{Advances in neural information processing systems}, 34:11054--11070.

\bibitem[{Popovi{\'c} and Belz(2021)}]{popovic2021reproduction}
Maja Popovi{\'c} and Anya Belz. 2021.
\newblock A reproduction study of an annotation-based human evaluation of mt outputs.
\newblock Association for Computational Linguistics (ACL).

\bibitem[{Prabhumoye et~al.(2023)Prabhumoye, Patwary, Shoeybi, and Catanzaro}]{Prabhumoye2023AddingID}
Shrimai Prabhumoye, Mostofa Patwary, Mohammad Shoeybi, and Bryan Catanzaro. 2023.
\newblock \href {https://api.semanticscholar.org/CorpusID:256868688} {Adding instructions during pretraining: Effective way of controlling toxicity in language models}.
\newblock In \emph{Conference of the European Chapter of the Association for Computational Linguistics}.

\bibitem[{Radfar et~al.(2020)Radfar, Shivaram, and Culotta}]{Radfar2020CharacterizingVI}
Bahar Radfar, K.~Shivaram, and Aron Culotta. 2020.
\newblock \href {https://api.semanticscholar.org/CorpusID:219561359} {Characterizing variation in toxic language by social context}.
\newblock In \emph{International Conference on Web and Social Media}.

\bibitem[{Rahman(2012)}]{rahman2012n}
Jacquelyn Rahman. 2012.
\newblock The n word: Its history and use in the african american community.
\newblock \emph{Journal of English Linguistics}, 40(2):137--171.

\bibitem[{Rast(2009)}]{Rast2009ContextAI}
Erich~H. Rast. 2009.
\newblock \href {https://api.semanticscholar.org/CorpusID:18494311} {Context and interpretation}.

\bibitem[{R{\"o}ttger et~al.(2021)R{\"o}ttger, Vidgen, Hovy, and Pierrehumbert}]{rottger2021two}
Paul R{\"o}ttger, Bertie Vidgen, Dirk Hovy, and Janet~B Pierrehumbert. 2021.
\newblock Two contrasting data annotation paradigms for subjective nlp tasks.
\newblock \emph{arXiv preprint arXiv:2112.07475}.

\bibitem[{Ruggeri et~al.(2023)Ruggeri, Antici, Galassi, Korre, Muti, and Barr{\'o}n-Cede{\~n}o}]{Ruggeri2023OnTD}
Federico Ruggeri, Francesco Antici, Andrea Galassi, Katerina Korre, Arianna Muti, and Alberto Barr{\'o}n-Cede{\~n}o. 2023.
\newblock \href {https://api.semanticscholar.org/CorpusID:258334802} {On the definition of prescriptive annotation guidelines for language-agnostic subjectivity detection}.
\newblock In \emph{Text2Story@ECIR}.

\bibitem[{Sahlgren and Lenci(2016)}]{sahlgren2016effects}
Magnus Sahlgren and Alessandro Lenci. 2016.
\newblock The effects of data size and frequency range on distributional semantic models.
\newblock \emph{arXiv preprint arXiv:1609.08293}.

\bibitem[{Sap et~al.(2019)Sap, Gabriel, Qin, Jurafsky, Smith, and Choi}]{sap2019social}
Maarten Sap, Saadia Gabriel, Lianhui Qin, Dan Jurafsky, Noah~A Smith, and Yejin Choi. 2019.
\newblock Social bias frames: Reasoning about social and power implications of language.
\newblock \emph{arXiv preprint arXiv:1911.03891}.

\bibitem[{Stokes and Cox(1970)}]{stokes1970aggressive}
Allen~W Stokes and Lois~M Cox. 1970.
\newblock Aggressive man and aggressive beast.
\newblock \emph{BioScience}, 20(20):1092--1095.

\bibitem[{Susanto et~al.(2024)Susanto, Wijanarko, Pratama, Hong, Idris, Aji, and Wijaya}]{susanto2024indotoxic2024demographicallyenricheddatasethate}
Lucky Susanto, Musa~Izzanardi Wijanarko, Prasetia~Anugrah Pratama, Traci Hong, Ika Idris, Alham~Fikri Aji, and Derry Wijaya. 2024.
\newblock \href {http://arxiv.org/abs/2406.19349} {Indotoxic2024: A demographically-enriched dataset of hate speech and toxicity types for indonesian language}.

\bibitem[{Tan and Celis(2019)}]{Tan2019AssessingSA}
Yi~Chern Tan and Elisa Celis. 2019.
\newblock \href {https://api.semanticscholar.org/CorpusID:202781363} {Assessing social and intersectional biases in contextualized word representations}.
\newblock \emph{ArXiv}, abs/1911.01485.

\bibitem[{Vargas et~al.(2021)Vargas, de~G{\'o}es, Carvalho, Benevenuto, and Pardo}]{vargas2021contextual}
Francielle Vargas, Fabiana~Rodrigues de~G{\'o}es, Isabelle Carvalho, Fabr{\'\i}cio Benevenuto, and Thiago Alexandre~Salgueiro Pardo. 2021.
\newblock Contextual-lexicon approach for abusive language detection.
\newblock \emph{arXiv preprint arXiv:2104.12265}.

\bibitem[{Yang et~al.(2021)Yang, Gan, Wang, Hu, Lu, Liu, and Wang}]{Yang2021AnES}
Zhengyuan Yang, Zhe Gan, Jianfeng Wang, Xiaowei Hu, Yumao Lu, Zicheng Liu, and Lijuan Wang. 2021.
\newblock \href {https://api.semanticscholar.org/CorpusID:237485500} {An empirical study of gpt-3 for few-shot knowledge-based vqa}.
\newblock \emph{ArXiv}, abs/2109.05014.

\bibitem[{Zampieri et~al.(2019)Zampieri, Malmasi, Nakov, Rosenthal, Farra, and Kumar}]{zampieri2019predicting}
Marcos Zampieri, Shervin Malmasi, Preslav Nakov, Sara Rosenthal, Noura Farra, and Ritesh Kumar. 2019.
\newblock Predicting the type and target of offensive posts in social media.
\newblock \emph{arXiv preprint arXiv:1902.09666}.

\bibitem[{Zec et~al.(2017)Zec, Soriani, Comoretto, and Baldi}]{zec2017suppl}
Slavica Zec, Nicola Soriani, Rosanna Comoretto, and Ileana Baldi. 2017.
\newblock Suppl-1, m5: high agreement and high prevalence: the paradox of cohen’s kappa.
\newblock \emph{The open nursing journal}, 11:211.

\end{thebibliography}
\bibliographystyle{acl_natbib}

\begin{table*}[h]
\begin{tabularx}{\textwidth}{lX}
\hline
\textbf{Term} & \textbf{Definition} \\
\hline
Aggression/Aggressiveness & Aggression in this context indicates hostile or rude attitudes, whether it involves readiness or not. \\
\hline
Aggressive & Being aggressive means showing hostile or rude attitudes, whether it involves readiness or not. \\
\hline
Offensiveness & General rudeness in a way that causes somebody to feel upset or annoyed because it shows a lack of respect. \\
\hline
Offensive & Being rude in a way that causes somebody to feel upset or annoyed because it shows a lack of respect. \\
\hline
External & Towards other people or parties. \\
\hline
Internal & Towards the self. \\
\hline
Construct & The mind-dependent object, namely ideas, perspectives, etc. \\
\hline
Inappropriate Language & Language uses that could have negative and unwanted impacts on people. \\
\hline
Biased Language & Biased Language contains obviously wrong or counterfactual expressions that target an individual or a group not limited to humans. \\
\hline
Offensive Language & Offensive Language shows intended aggressiveness toward others. \\
\hline
Hate Speech & Hate Speech is an offensive language of intense external aggressive intention with explicit targets rooted in explicit or implicit false constructs. \\
\hline
\end{tabularx}

\caption{Definitions of Terms}
\label{tab:definitions}
\end{table*}

\appendix

\section{Annotator Codebook}
\label{code}
\subsection{General Deﬁnitions}
A list of short-cut definitions is presented in Table \ref{tab:definitions}. Please see the methodology for further validations.

\subsection{Annotation Instruction for two Indicators}
\textbf{Aggression} will be assessed regarding every distinct negative, rude, or hostile attitude. Please see Table \ref{aggression} and general description below for more information about speciﬁc language use. Computation logic: If the score is less or equal to 1, the aggression level will be 1. If the score exceeds 1, the aggression level will be 2. Otherwise, the aggression level will be 0.
\begin{itemize}
    \item Level refers to the general linguistic category of each item.
    \item Item name includes the names of aggression-related items.
    \item Category refers to the category that indicates how the item is related to aggression.
    \begin{itemize}
        \item Aggressive items / AI (1 point): are aggressive by themselves.
        \item Aggression catalyzers / AC (.5 point): are unaggressive themselves and function to boost the aggressive level.
        \item Expressions from the same item category only count once; for example, if there are two different aggressive noun phrases, the score will be one rather than two.
        \item Override Rule: The overall relative aggression score will be 0 if there is no aggressive item.
        \item SPECIAL CASE: False constructs are non-aggressive. But when people pair false constructs with other aggressive catalyzers, they become aggressive items (but with .5 point) and should be seen as aggression bases. For example, how come your people really believe in flat earth?
    \end{itemize}
    \item Example contains examples of each item.
\end{itemize}
\textbf{Direction of Language Intent} (External or Non-external) evaluates Whether the language targets other(s) explicitly. The direction is decided regarding the direction of aggression, which means even statements about speakers’ selves could contain aggression against others.

\section{Annotator Surveys}
\label{anno}
\subsection*{Specialties}
\begin{itemize}
    \item Annotator 1 without criteria: Internet Marketing \& Data Analytics
    \item Annotator 2 without criteria: Corpus Linguistics \& Syntax
    \item Annotator 1 with criteria: Semantics Analysis \& Syntax \& Corpus Linguistics
    \item Annotator 2 with criteria: Socio-linguistics \& Language Acquisition
\end{itemize}

\subsection*{Aside from mainstream English, are you familiar with any regional dialects, sociolects, or linguistic styles more common in minority communities and groups?}
\begin{itemize}
    \item Annotator 1 without criteria: Yes
    \item Annotator 2 without criteria: Yes
    \item Annotator 1 with criteria: Yes
    \item Annotator 2 with criteria: Yes
\end{itemize}

\subsection*{Approximately how many hours did it take you to complete all the annotations assigned to you?}
\begin{itemize}
    \item Annotator 1 without criteria: 4
    \item Annotator 2 without criteria: 4.5
    \item Annotator 1 with criteria: 5 (criteria-based training) + 7 (annotation)
    \item Annotator 2 with criteria: 5 (criteria-based training) + 8 (annotation)
\end{itemize}

\subsection*{How confident are you in the accuracy of the annotations you completed? (1-5)}
\begin{itemize}
    \item Annotator 1 without criteria: 2. No so confident, many African American English I found hard to understand accurately
    \item Annotator 2 without criteria: 3. I am confident about my annotations identifying explicit toxic expressions and hate speech, but less confident in others.
    \item Annotator 1 with criteria: 4.5. I'm pretty confident, though I'm not an African American English native speaker. I studied AAE corpus before, so I consider myself familiar with AAE. About that DI, sometimes I think it could go either way cause their tweets ain't just one sentence. For AG, the score generally matches what I think about aggression. All in all, this dataset is easier than the one with political stuff. I don't know too much about politics.
    \item Annotator 2 with criteria: 4. Yes, I think AAE is not really an issue. The AG scoring guide helps break things down to the word level. Basically, it doesn't really matter if the phrases are used differently or not; as long as they are seen as aggressive by some people, they'll be taken as aggressive. But it really takes a lot of time and effort just to highlight each aggressive item and categorize the aggression. DI seemed pretty straightforward to me at first, but after our group discussion, I realized there could also be other interpretations.

\end{itemize}

\subsection*{Looking back at your annotations after a month has passed, how did you feel about the quality and accuracy of the work you originally completed?}
\begin{itemize}
    \item Annotator 1 without criteria: Still confused about many tweets.
    \item Annotator 2 without criteria: There could be different interpretations. It's really about the larger context. 
    \item Annotator 1 with criteria: Not really much in terms of toxicity. DI's still kinda confusing in a couple of cases.
    \item Annotator 2 with criteria: Basically the same as when I finished it up
\end{itemize}

\section{Confusion Matrices (Figure 2-5)}
\label{sec:appendix}

\centering
\begin{figure}[h]
    \includegraphics[width=\linewidth]{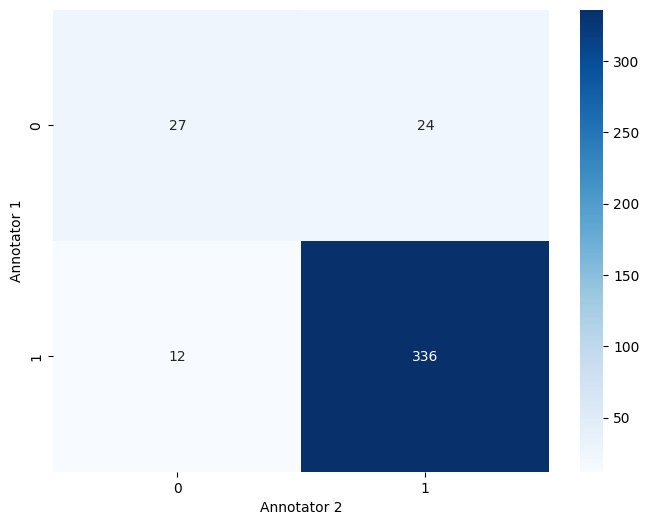}
    \caption{Confusion Matrix on Direction Intent Annotation}
\end{figure}

\begin{figure}[h]
    \includegraphics[width=\linewidth]{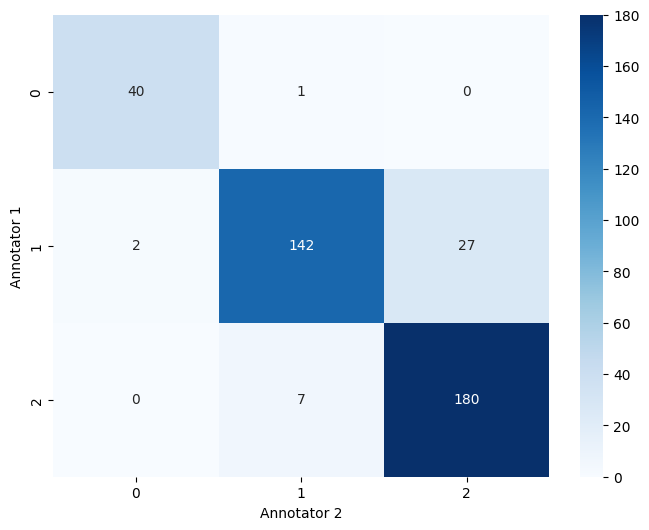}
    \caption{Confusion Matrix on Aggression Annotation}
    \centering
\end{figure}

\begin{figure}[h]
    \includegraphics[width=\linewidth]{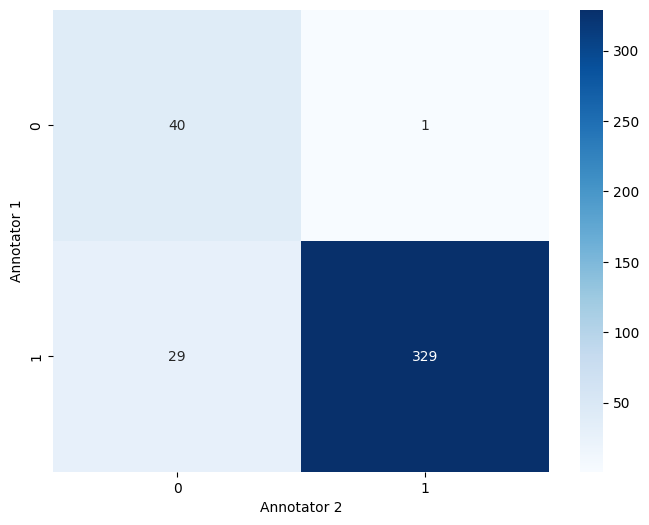}
    \caption{Confusion Matrix on Toxicity Annotation with Criteria}
\end{figure}

\begin{figure}[h]
    \includegraphics[width=\linewidth]{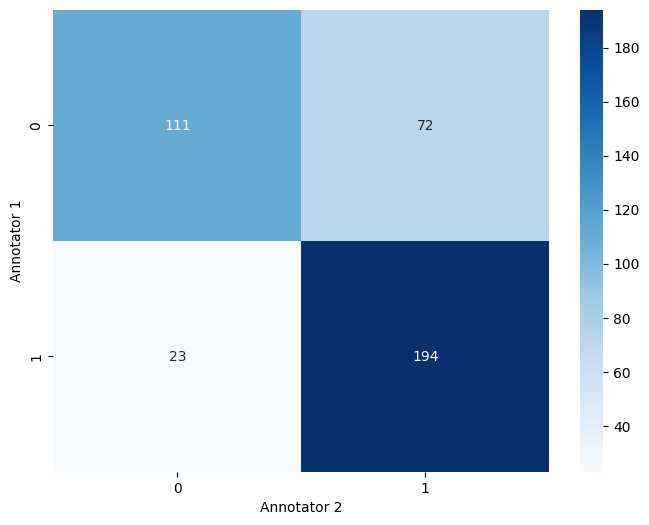}
    \centering
    \caption{Confusion Matrix on Toxicity Annotation without Criteria}
\end{figure}

\section{Annotation Distribution (Figure 6-9)}
\label{sec:appendix2}

\centering
\begin{figure}[h]
    \includegraphics[width=\linewidth]{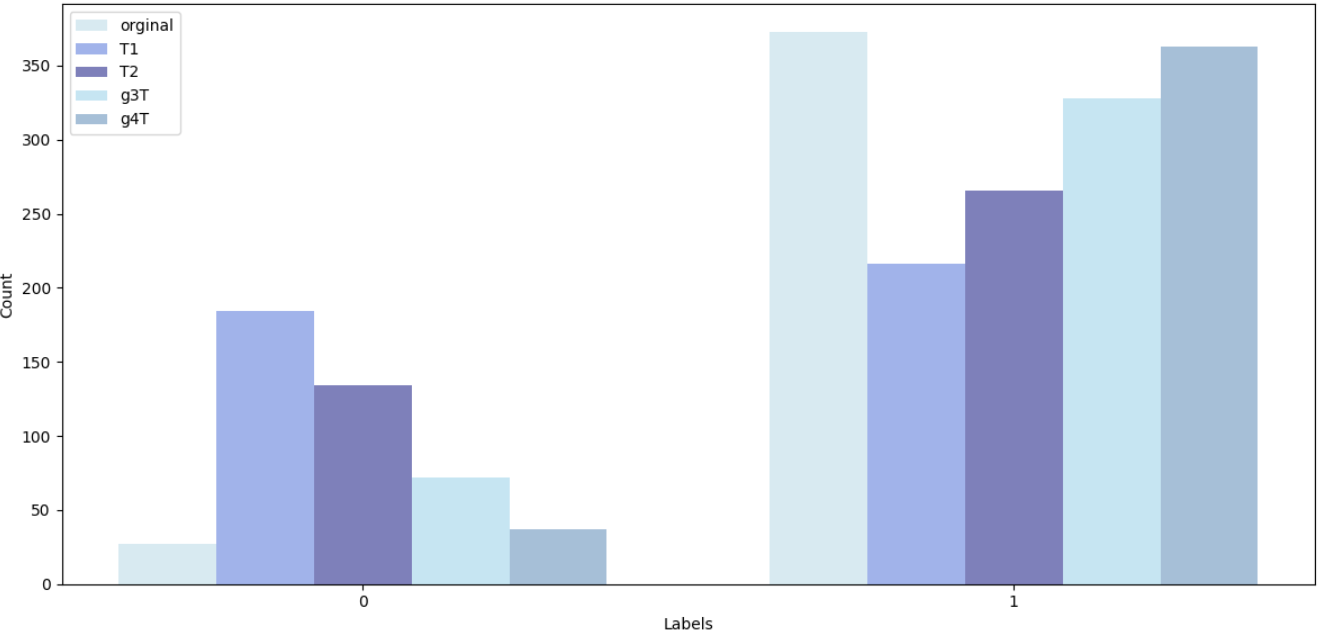}
    \caption{Distribution of Toxicity Annotation without Criteria}
\end{figure}

\begin{figure}[h]
    \includegraphics[width=\linewidth]{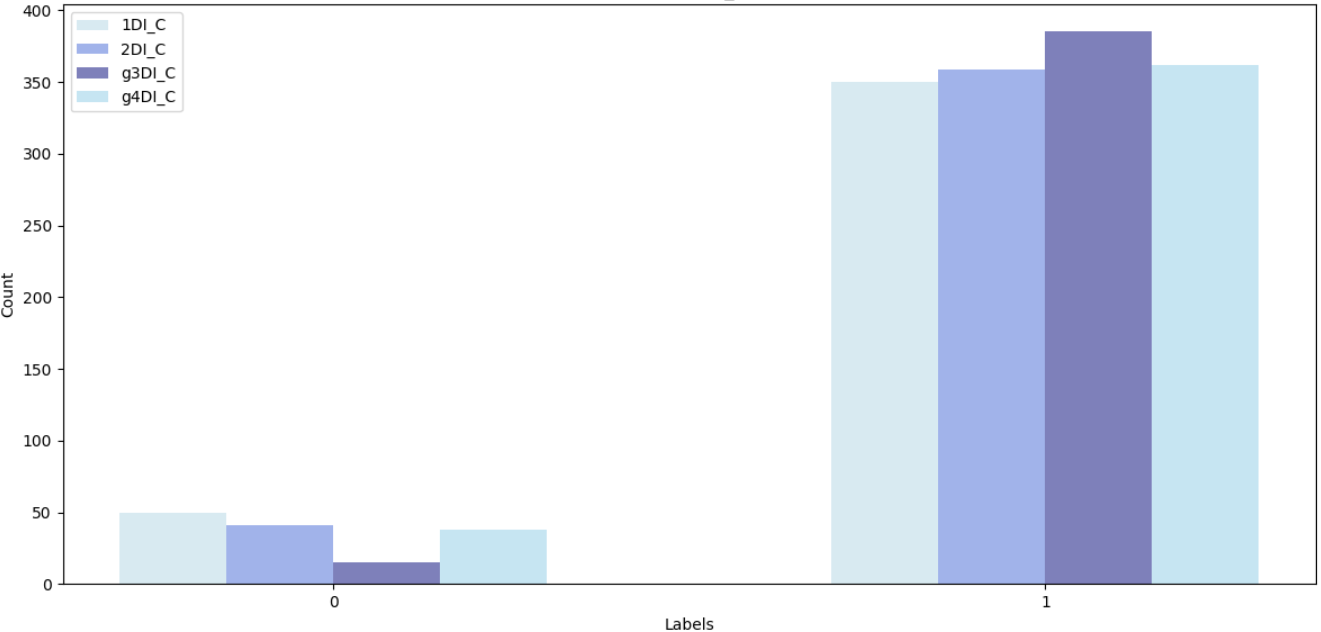}
    \caption{Distribution of Direction of Language Intent Annotation with Criteria}
    \centering
\end{figure}

\begin{figure}[h]
    \includegraphics[width=\linewidth]{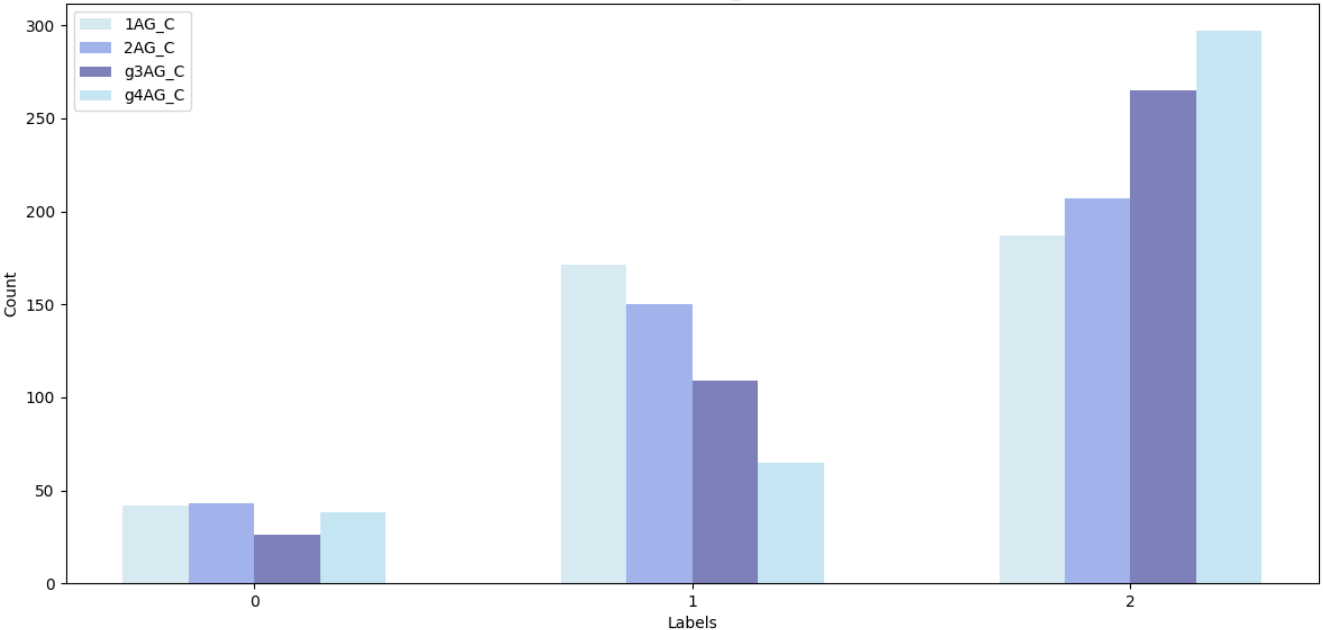}
    \caption{Distribution of Aggressive Level Annotation with Criteria}
\end{figure}

\begin{figure}[h]
    \includegraphics[width=\linewidth]{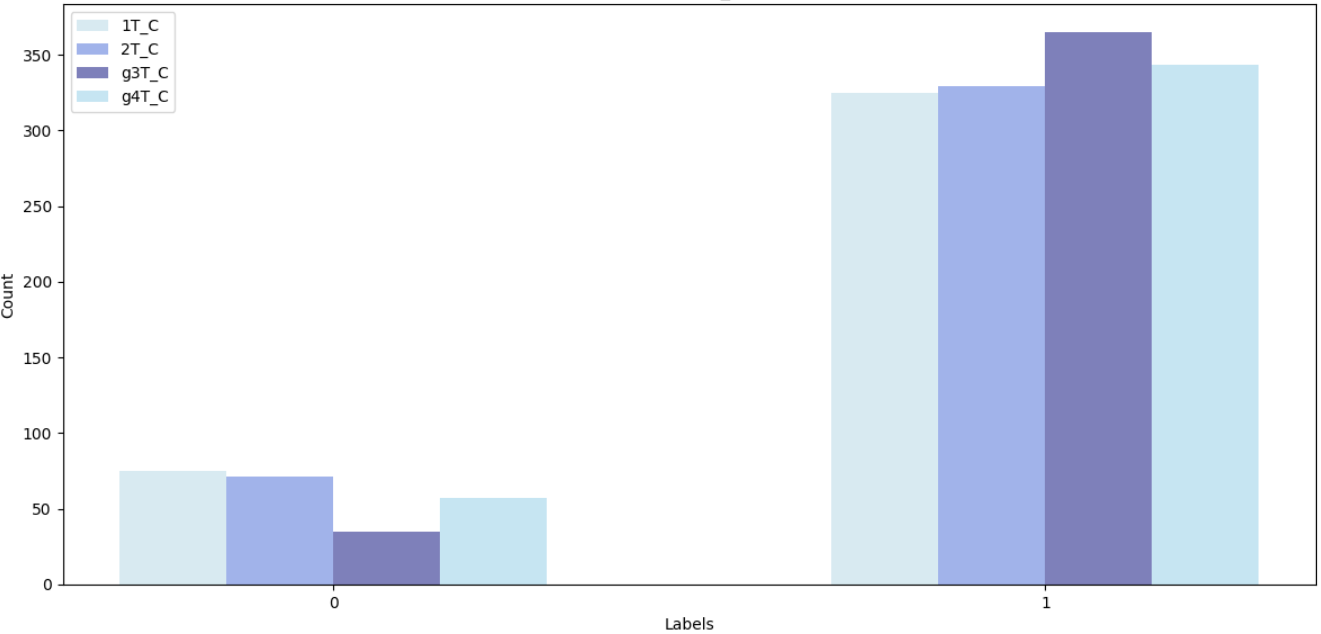}
    \centering
    \caption{Distribution of Toxicity Annotation with Criteria}
\end{figure}

\end{document}